\documentclass[journal,10pt]{IEEEtran}
\usepackage{times}
\usepackage{epsfig}
\usepackage{graphicx}
\usepackage{amsmath}
\usepackage{amssymb}
\usepackage{subcaption,float}
\usepackage{color}

\newcommand{\wbw}[1]{#1}
\newcommand{\revision}[1]{#1}
\begin{document}
\title{Image Comes Dancing with Collaborative Parsing-Flow Video Synthesis}

\author{Bowen Wu$^*$, Zhenyu Xie$^*$, Xiaodan Liang$^\dagger$, Yubei Xiao, Haoye Dong, Liang Lin% <-this % stops a space
\thanks{$^*$ Bowen Wu and Zhenyu Xie contribute equally to this work}

\thanks{$^\dagger$ Corresponding author: Xiaodan Liang, E-mail:xdliang328@gmail.com}

\thanks{All authors are with Sun Yat-sen University, China. Liang Lin is also with Dark Matter AI Research, China.}
}


\markboth{IEEE TRANSACTIONS ON IMAGE PROCESSING,~Vol.~xx, No.~x, xxxxxx~xxxx}%
{Shell \MakeLowercase{\textit{et al.}}: Bare Demo of IEEEtran.cls for IEEE Journals}



\maketitle
\begin{figure*}[h]
    \centering
    \includegraphics[width=1.0\linewidth]{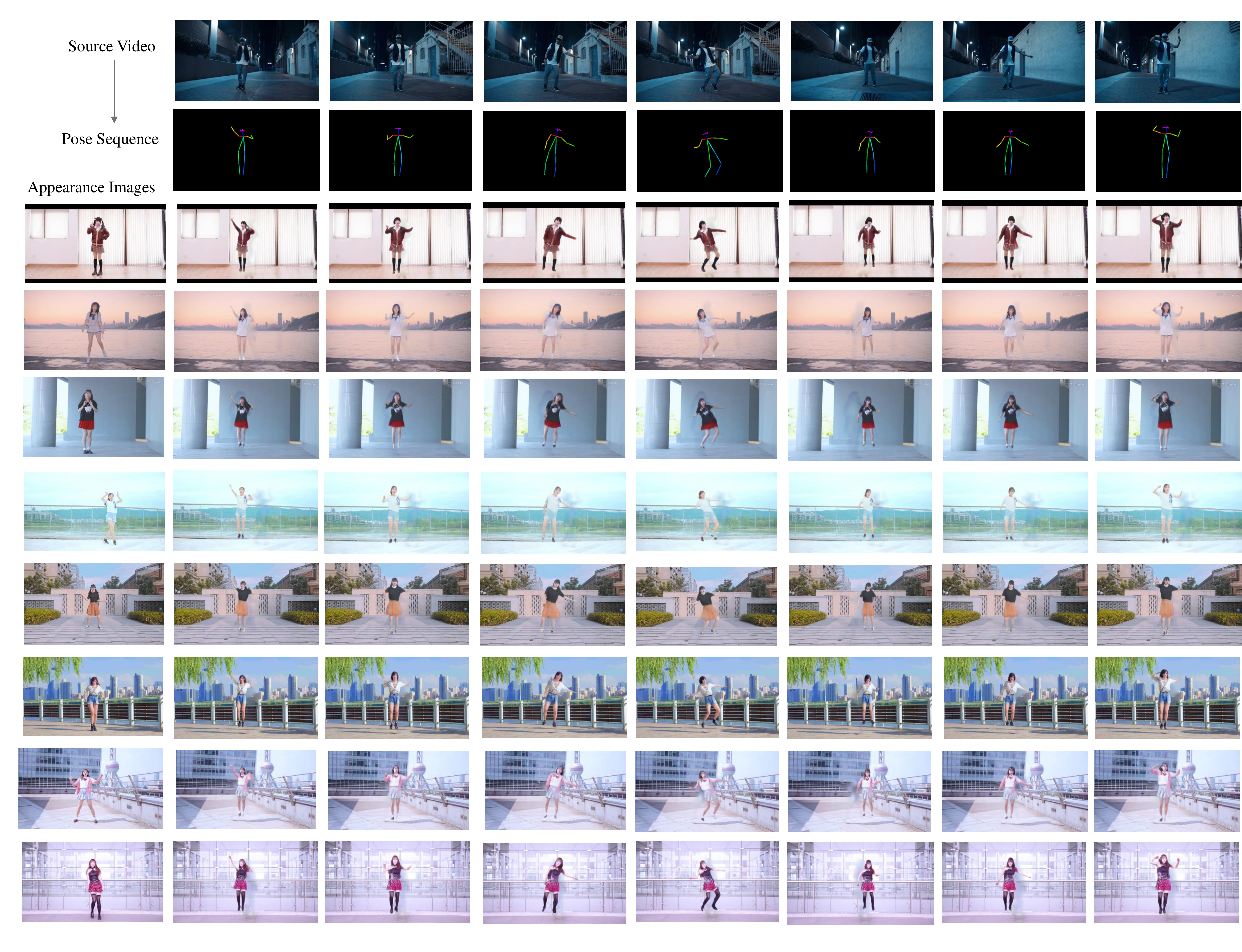}
    \caption{\textbf{Synthesized Videos of Our Approach.} Given an appearance image of a target person and a video of a source person, our model generates a video of the target person which retains the source video motions. The first column is the appearance images of three target persons. The first two rows are the source video and its estimated pose sequence, respectively. The next three rows are the synthesized video frames via both retaining the target person's appearance and the source video motions.}
    \label{fig:fig1}
\end{figure*}
% \twocolumn[{%
% \renewcommand\twocolumn[1][]{#1}%
% \maketitle
%     \includegraphics[width=0.97\linewidth]{fig/firstpage.pdf}
%     \captionof{figure}{\textbf{Synthesized Videos of Our Approach.} Given an appearance image of a target person and a video of a source person, our model generates a video of the target person which retains the source video motions. The first column is the appearance images of three target persons. The first two rows are the source video and its estimated pose sequence, respectively. The next three rows are the synthesized video frames via both retaining the target person's appearance and the source video motions.\newline\newline}
%     \label{fig:fig1}
% }]

\begin{abstract}
Transferring human motion from a source to a target person poses great potential in computer vision and graphics applications. 
A crucial step is to manipulate sequential future motion while retaining the appearance characteristic.
Previous work has either relied on crafted 3D human models or trained a separate model specifically for each target person, which is not scalable in practice.
This work studies a more general setting, in which we aim to learn a \emph{single} model to parsimoniously transfer motion from a source video to any target person given only one image of the person, named as Collaborative Parsing-Flow Network (CPF-Net). The paucity of information regarding the target person makes the task particularly challenging to faithfully preserve the appearance in varying designated poses.
To address this issue, CPF-Net integrates the structured human parsing and appearance flow to guide the realistic foreground synthesis which is merged into the background by a spatio-temporal fusion module.
In particular, CPF-Net decouples the problem into stages of human parsing sequence generation, foreground sequence generation and final video generation. The human parsing generation stage captures both the pose and the body structure of the target. The appearance flow is beneficial to keep details in synthesized frames. The integration of human parsing and appearance flow effectively guides the generation of video frames with realistic appearance. Finally, the dedicated designed fusion network ensure the temporal coherence. We further collect a large set of human dancing videos to push forward this research field. Both quantitative and qualitative results show our method substantially improves over previous approaches and is able to generate appealing and photo-realistic target videos given any input person image. All source code and dataset will be released at https://github.com/xiezhy6/CPF-Net.
\end{abstract}


\section{Introduction}
Transferring human motion from one subject to another target subject is a long-standing challenge in computer vision and graphics. Such techniques enable us to generate animations and videos of a human with realistic motion and appearance, and can be particularly useful in virtual reality, video games, art creation, and so forth. A crucial part of the problem is modeling the characteristics (e.g., body shape, clothing appearance) of the target subject. Previous research has studied settings of using 3D human models to this end~\cite{cheung2004markerless,hecker2008real,shen2015automatic,villegas2018neural}. Recent work~\cite{chan2018dance} instead learns a personalized neural network from videos of each target subject, which is then used to synthesize new videos of arbitrary motions of the particular target person.
Despite the impressive results, however, both the 3D character models and the subject-specific neural models can hardly scale up in practice. For example, consider deploying an application of creating dancing videos for any users on social media, it is computationally and economically prohibitive to train a model for each user.

In this work, we consider a more practical setting of using a \emph{single} model to transfer motions to any target person based on only \emph{one image} of the person. The image provides the appearance characteristics of the target subject. With this image and a video of source subject, we aim to synthesize a realistic video wherein the target performs the same motions as in the source video. Figure~\ref{fig:fig1} illustrates our task.
Applying the same model for everybody and requiring only a single target image can enable very scalable applications, simultaneously also imposes unique technical challenges in generating photo-realistic videos with faithful target appearance and smooth designated motions.
%in order to smoothly synthesize the designated motions, faithfully preserve the target appearance, and in the end yield a high-quality photo-realistic video.
% A different line of research~\cite{YangPoseGuid,Ma2017PoseGP,Si_2018_CVPR,Cai_2018_ECCV,Lakhal2018PoseGH,Pumarola_2018_CVPR} explored the problem of generating a video from a subject's image with predicted motions of the same person. The motion transfer task we study here is distinct in that we adapt the motions of a different source subject to the target person, which allows fine-grained motion control and requires dealing with diverse properties of 
% the source and target, such as appearance, body proportions, video background, and others.

In contrast to previous settings where videos of a target subject provide rich person information and end-to-end neural models can straightforwardly be applied~\cite{wang2018vid2vid}, a solution to this task must ensure accurate modeling of the target subject despite the paucity of information from the single image, and the model must be applicable to any given target subject without specific training. This necessitates a structured method that incorporates the human body and motion structures for effective motion transferring.

To address the above mentioned challenges, we propose a Collaborative Parsing-Flow method (CPF-Net) that integrates structured human parsing and appearance flow to guide the video synthesis. Specifically, a human parsing, as a semantic segmentation of human parts (e.g., head, legs, etc)~\cite{Liang2018look,liang2016semantic,liang2015human,zhou2018adaptive,gong2018instance}, effectively captures both the pose and the structure of the target person, which can not only guide to generate video frames of the target in various poses but guide the generator to synthesize different body part precisely.
% and guides to generate video frames of the target in various poses with faithful appearance. 
The appearance flow, indicating the pixel corresponding relation between the source and target image, can be used to warp the source appearance to the target shape and keep the source image appearance details as much as possible. 
%With the integration of human parsing and appearance flow, our method 
In particular, we decompose the problem into three stages, namely, pose-guided human parsing synthesis, parsing-flow-guided foreground synthesis, and spatio-temporal fusion-based video synthesis. In the first stage, we generate a sequence of the target's human parsing conditioning on the pose of the source video and the appearance parsing of the target person. 
In the second stage, we first utilize a flow regression network to estimate a appearance flow between the appearance pose and each source video pose, which is used to warp the extracted feature of the appearance foreground. Collaborating the human paring and warped appearance feature, the target foreground is generated. In the third stage, the foreground and background are fused together while keeping the temporal smoothness by taking the previous synthesized frame as the input of the fusion network.

%the human parsing results are fused with the target's appearance image to synthesize photo-realistic frames in the output video. The decoupling of human parsing sequence generation and final video synthesis further enables us to apply off-the-shelf video-to-video techniques~\cite{Liu2017surveillance, zhu2017flow,saito2017temporal, tulyakov2018mocogan,wang2018vid2vid} with optical flow on both stages, respectively, in order to ensure temporal smoothness.

To push the limit of the arbitrary human motion transfer research field, we further build a large collection of dancing videos. The dataset consists of 428 videos of different people performing various dances in diverse scenes. 
We conduct extensive experiments to compare with strong baselines adapted from previous approaches~\cite{wang2018pix2pixHD,wang2018vid2vid,soft-gated}. Both quantitative and qualitative results demonstrate the superiority of the proposed method. Figure~\ref{fig:fig3} and Figure~\ref{fig:fig4} show our method gains much better visual quality compared with existing baseline methods.

\section{Related Work}
\subsection{Motion Transfer}
% \noindent\textbf{.}
% Transferring motions from a source to any target subject is a long-standing problem~\cite{cheung2004markerless,hecker2008real,shen2015automatic,villegas2018neural,chan2018dance,Liu2018NeuralAA,Shysheya2019TexturedNA}, widely used in virtual reality, video games, art creation, and so forth. The representation of motions from source subject is generally based on pose sequence while the model of characteristics can be a 3d character model~\cite{cheung2004markerless,hecker2008real,shen2015automatic,villegas2018neural} and a motions video~\cite{chan2018dance} of the target person. Transferring motions to a 3d character model are classic and less difficult task as the process of video rendering can be done by the game engine with predefined 3d model.
Extensive works~\cite{cheung2004markerless,hecker2008real,shen2015automatic,villegas2018neural,chan2018dance,Liu2018NeuralAA,Shysheya2019TexturedNA} of motion transfer have been done in the past decades and are widely used in various areas. Liu \textit{et al.}~\cite{Liu2018NeuralAA} has explored to build a motion transfer system based on a rough 3D human model. Villegas \textit{et al.} attempted to transfer motions from a source to any 3D target characters with forward kinematics and inverse kinematics. 

All of the above methods have achieved positive results but they must be supported by 3D models which are difficult to be acquired and complex. Instead, our proposed CPF-Net is free of fine-grained 3D character models and only requires one image of the target person to transfer motions from source video to the target person. Breaking through the limitations of 3D models, a very recent work~\cite{chan2018dance} proposed a framework based on pix2pixHD~\cite{pix2pix2017} to straightforwardly train an end-to-end neural network that can transform 2D pose to the target person. However, it needs to train a personalized model for each target person. 
Opposed to it, another family of methods called image animation~\cite{Siarohin2019FirstOM, Siarohin2019AnimatingAO, Tulyakov2018MoCoGANDM} could animate any source object according to the motion of a driving video in single model. 
MoCoGAN~\cite{Tulyakov2018MoCoGANDM} maps a sequence of video frames to a sequence of random vectors. Motion transfer can be realized by keeping the content part in the random vector while changing the motion part in the random vector. However, this method can only synthesize low resolution videos without appearance details.
~\cite{Siarohin2019FirstOM,Siarohin2019AnimatingAO} utilize the keypoint detector to extract the keypoints of the source image and the driving frame, which are used to estimate the optical flow between the source image and the driving frame. However, since these methods train the keypoint detector, dense motion network, and generator network in an end-to-end manner and there is no supervision for the keypoint detector,
these methods often fail on complex human motions.
Our approach can synthesize high resolution videos (1280x720) and handle complex motions, which is superior of these image animation methods.

\subsection{Image Synthesis}
Since there has been much progress in Generative Adversarial Networks (GAN)~\cite{2014arXiv1406.2661G}, image synthesis has achieved incredible results. Some works on image synthesis such as BigGAN \cite{Brock2018LargeSG} generate pictures that even humans can not distinguish them from the real images. Image synthesis can be generally classified into two categories, unconditional synthesis, and conditional synthesis. For unconditional synthesis, some work based on GAN~\cite{2014arXiv1406.2661G} or VAE~\cite{kingma2013auto} learns a mapping function from noise to real image. For conditional image synthesis, conditional models usually extending Conditional Generative Adversarial Nets synthesize images given either class category, human pose, scene graphs, and another image~\cite{ma2017pose,johnson2018image,shrivastava2017learning}. Conditional Generative Adversarial Nets (C-GAN)~\cite{Mirza2014ConditionalGA} is a variant of GAN, generating images with given desired attributes by embedding conditional information on both generator and discriminator. Based on C-GAN, some works achieved an impressive result on image-to-image translation ~\cite{pix2pix2017,wang2018pix2pixHD}. Face generation plays an important role in human image synthesis. Many works are focusing on face synthesis, such as Tero Karras \textit{et al.}~\cite{Karras2017ProgressiveGO}, which have gained incredible results on face synthesis. Guha \textit{et al.} \cite{Balakrishnan2018SynthesizingIO} proposed a method to synthesize images of humans in unseen poses. Their method could synthesis a person's image of the desired pose via the given source image and target pose.

\subsection{Video Synthesis}
Many works on specific problems of video synthesis, including video matting and blending~\cite{Bai2009VideoSR,Chen2013MotionAwareGD}, video super-resolution~\cite{Shi2016RealTimeSI} and video inpainting~\cite{Wexler2007SpaceTimeCO} can be applied to other general video synthesis tasks.
%There are many works of video synthesis, including video matting and blending~\cite{Bai2009VideoSR,Chen2013MotionAwareGD}, video super-resolution~\cite{Shi2016RealTimeSI} and video inpainting~\cite{Wexler2007SpaceTimeCO}. These tasks above can be considered as some specific problems of video synthesis and their proposed method can be easily applied to other general video synthesis tasks. 
Among them, vid2vid~\cite{wang2018vid2vid} is a video-to-video translation framework and is capable of generating incredible high-resolution videos. Though it is possible to simply adapt this network to additionally condition on the target appearance image for our task, we show in our empirical study (section~\ref{sec:exp}) that such a simple solution fails to produce satisfactory results due to the diversity of target persons and the paucity of information provided by the single images of targets. 
\revision{Few-shot vid2vid~\cite{Wang2019FewshotVS} is the improved version of vid2vid, which aims at synthesizing video for an arbitrary person through a single model. 
Apart from the pose sequence, few-shot vid2vid requires some example images for the model, which can provide the appearance information of the target person.
During training, it trains the vid2vid model in a few-shot paradigm. During testing, to synthesize video for a new person that does not exist in the training set, few-shot vid2vid only needs a pose sequence and few example images as the inputs of the model. Different from the above two methods, Liquid Warping GAN~\cite{Liu2019LiquidWG} resorts to the transformation flow between the source person and the reference person, which is used to align the feature of the source person to the target pose. The transformation flow can be obtained according to the SMPL~\cite{Loper2015SMPLAS} models of the source person and the reference person.
Compared with the existing methods, our method incorporates the structured model and appearance flow to enable efficient structure learning and good generalization capability.}

\begin{figure*}[ht]
    \centering
    \includegraphics[width=0.9\linewidth]{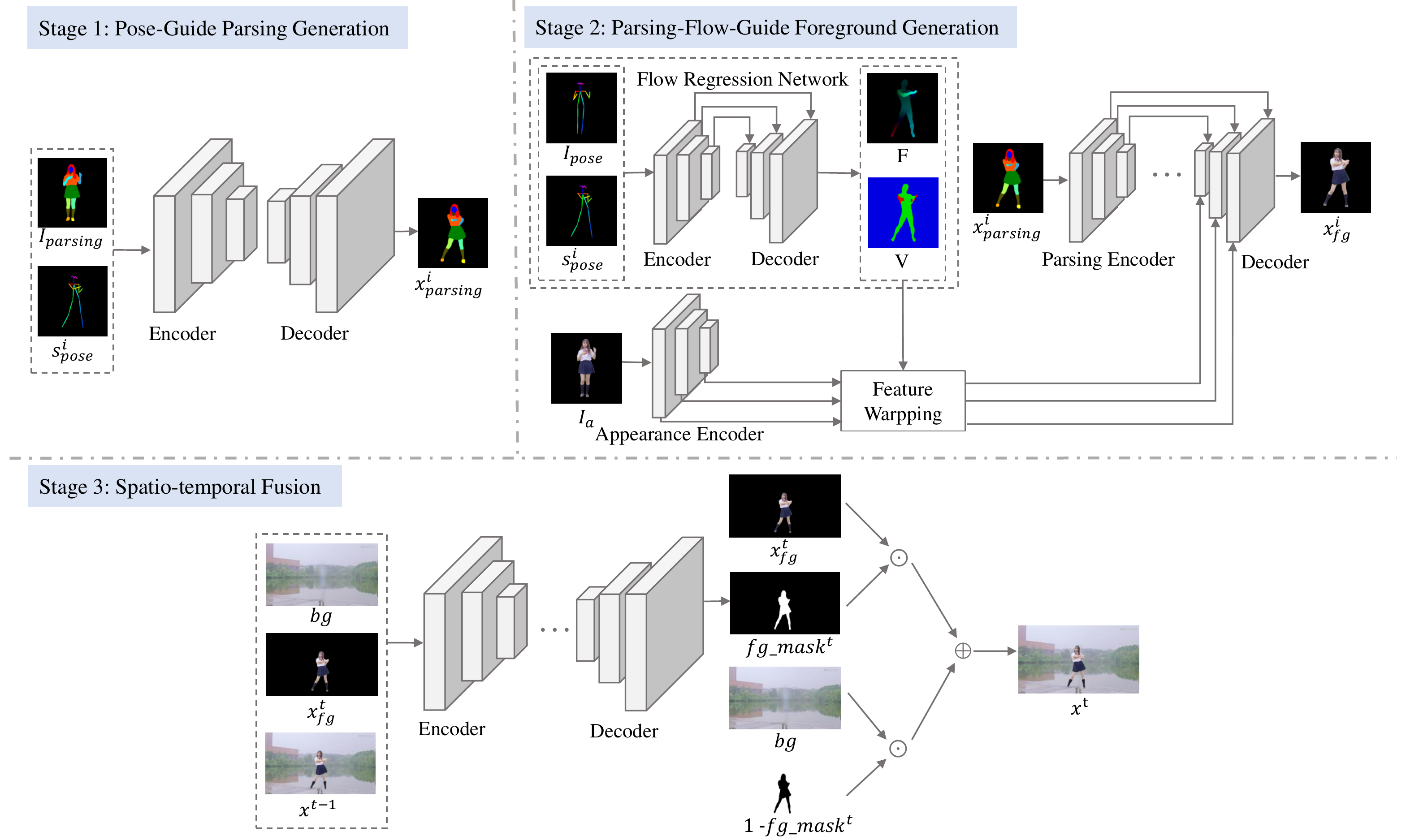}
    \caption{\textbf{Overall architecture of the proposed CPF-Net.} In the first stage, we feed the appearance parsing $I_{parsing}$ and source video pose $s_{pose}^i$ into pose-guide parsing generation module to obtain the target parsing $x_{parsing}^i$. In the second stage, appearance pose $I_{pose}$ together with $s_{pose}^i$ are fed into the flow regression network to obtain the appearance flow map $F$ and visibility map $V$. Then the appearance foreground $I_a$, $x_{parsing}^i$, $F$ and $V$ are fed into parsing-flow-guide foreground generation module to synthesize the target foreground $x_{fg}^i$. In the third stage, we first extract the background from the appearance image and inpaint it as $bg$. Then, the inpainted background $bg$, current synthesized foreground $x^t_{fg}$ and previous synthesized frame $x^{t-1}$ are fed into spatio-temporal fusion network to obtain current foreground mask $fg\_mask^t$, which is used to fuse $x^t_{fg}$ and $bg$ to synthesize the current target frame $x^t$.}
    \label{fig:framework}
\end{figure*}

\section{CPF-Net}\label{sec:method}
% Different from our new setting, previous motion transfer works~\cite{wang2018vid2vid,chan2018dance} train a separate model specific to each target person, where end-to-end networks can straightforwardly be applied to fit videos of the target. Though it is possible to simply adapt the networks to additionally condition on the target appearance image for our task, we show in our empirical study (section~\ref{sec:exp}) that such a simple solution fails to produce satisfactory results due to the diversity of target persons and the paucity of information provided by the single images of targets. This raises the necessity of introducing structured modeling (e.g., for human body and appearance) to enable efficient structure learning and the good generalization capability.

% To address the above mentioned challenges, we propose a Collaborative Parsing-Flow model (CPF-Net) that integrates structured human parsing and appearance flow to guide the video synthesis.
Collaborative Parsing-Flow model (CPF-Net), integrating structured human parsing and appearance flow to guide the video synthesis, was proposed to tackle the challenges in human motion transfer.
% A human parsing segments human parts such as head and legs and provides an effective expression of the target person structure.
The human parsing is an effective tool for expressing the target person structure, which segments person into different parts, such as the head part and legs part.
%The augmentation of human parsing modeling~\xyb{modelling to modeling} enables accurate reconstruction of the target appearance in varying motions. 
Inspired by~\cite{intrinsic-flow}, we utilize the appearance flow generated from a flow regression network to guide the synthesis of the target frame which retains most details of the appearance frame. Moreover, to fuse the foreground and background properly and keep the temporal smoothness among consecutive frames, we propose a spatio-temporal fusion network. Benefit from the collaborative structural learning guided by human parsing and flow, our approach is capable of producing photo-realistic and spatio-temporal coherent dancing videos of any target input. Figure~\ref{fig:framework} illustrates the overall architecture of our approach. We decompose the modeling into three stages which will be fully elaborated in section~\ref{sec:pose-guide}, section~\ref{sec:parsing-flow-guide} and section~\ref{sec:spatio-temporal} respectively.

% We first formally define our task and establish the notations in section~\ref{sec:task}, and present our proposed approach in section~\ref{sec:pose-guide}, section~\ref{sec:parsing-flow-guide} and section~\ref{sec:spatio-temporal} respectively.

\subsection{Problem Formulation and Notation}\label{sec:task}
We consider the problem of making a target person in a static image dance like the person in a source video, by transferring motion from the source to the target. Formally, let $\mathbf{S}=\{s^1, s^2, \dots, s^T\}$ be a sequence of frames in the source dancing video, and $I_a$ is an appearance image of the target person. Given the input pair $(S, I_a)$, we aim to generate a new video $\mathbf{X}=\{x^1, x^2, \dots, x^T\}$ wherein the target person performs the same dance motions as the source person. In particular, a model to the task $P(\mathbf{X} | \mathbf{S}, I_a)$ should generally apply to arbitrary source dancing video and any target person, and able to synthesize a target video that smoothly retains the motions of the source and faithfully preserve the appearance of the target.

\subsection{Pose-Guide Parsing Generation}\label{sec:pose-guide}
The first stage of CPF-Net fuses source motion with the target human structure as an intermediate representation of motion transfer. Specifically, we first apply a pre-trained pose detector on the source video frames $\mathbf{S}$ to extract the pose sequence $\mathbf{S_{pose}}$, and apply a pre-trained human parser on the target appearance image $I_a$ to obtain the human parsing map $I_{parsing}$.

\noindent\textbf{Pose Sequence Smoothing.} The predicted pose sequence $\mathbf{S_{pose}}$ is inevitable with some noise and these noise will cause temporal inconsistency on generated video. Therefore, Savitzky–Golay filter~\cite{Steinier1964SmoothingAD} is employed to smooth predicted pose sequence.

The goal of this stage is to generate the human parsing sequence $\mathbf{X_{parsing}}$ of the target parsing $I_{parsing}$ that implements the source pose sequence $\mathbf{S_{pose}}$:
% The goal of this stage is to transfer the source pose sequence $\mathbf{S_{pose}}$ to the target parsing map $I_{parsing}$ thus generate the human parsing sequence $\mathbf{X_{parsing}}$:
\begin{equation}
\mathbf{X_{parsing}} = G_{parsing}(\mathbf{S_{pose}}, I_{parsing}),
\label{eq:parsing_gan}
\end{equation}
where $G_{parsing}$ is a pose-guided parsing generation model. 

%More specifically, each frame of the pose sequence is joint with the target initial parsing $I_{parsing}$ and fed into the model to generate a preliminary parsing map.

% Similar to~\cite{soft-gated}, we utilize the ResNet-like generator which has 9 residual blocks. The generator takes the target appearance parsing $I_{parsing}$ and pose of source video frame $s_{pose}^i$ as inputs, and output the target parsing $x_{parsing}^i$ in that pose. 
 
 To train the parsing generation model, we construct the training data from a set of human dancing videos. Concretely, we treat each video as the source video $\mathbf{S}$, and the source subject in the video as the target person as well. A particular frame in the video is selected as the target appearance image $I_a$ (the selection criteria is described in Section~\ref{sec:dataset}). In this way, the human parsing map of $I_a$ can serve as $I_{parsing}$ of this stage and predicted parsing maps of source video frames can serve as ground true while training.
 % In this way, the human parsing maps of the source video frames by the pre-trained human parser can serve as the ground truth for model training in this stage.
% To train the parsing generation model, we first predict human parsing maps of each dancing videos and each appearance image $I_a$(refer to sec~\ref{sec:dataset} for more details).
 
 Similar to~\cite{soft-gated}, we utilize the ResNet-like generator which has 9 residual blocks. The generator takes the target appearance parsing $I_{parsing}$ and pose of source video frame $s_{pose}^i$ as inputs, and outputs the target parsing $x_{parsing}^i$ in that pose. As for loss functions, we utilize $\mathcal{L}_1$ loss between the generated parsing maps and ground truth parsing maps. Besides, we use a pixel-wise softmax loss $\mathcal{L}_{par}$ from LIP~\cite{2017lip} to improve the quality of results. Let $x_{p}' \in R^{W*H*C}$ and $x_{p} \in R^{W*H*C}$ be the predicted parsing result and parsing ground truth, where $W$ and $H$ denote the width and height of input image respectively and $C$ is the number of semantic labels. Then $\mathcal{L}_1$ and $\mathcal{L}_{par}$ can be formulated as below:
 \begin{equation}
 \label{eq:loss_l1}
 \begin{aligned}
     \mathcal{L}_1 &= \sum_{w=0}^W\sum_{h=0}^H\sum_{l=0}^C|x_{p}(w,h,l) - x_{p}'(w,h,l)|, \\
     \mathcal{L}_{par}&=-\sum_{w=0}^W\sum_{h=0}^H\sum_{l=0}^Cx_{p}(w,h,l)log[x_{p}'(w,h,l)].
 \end{aligned}    
 \end{equation}

\subsection{Parsing-Flow-Guide Foreground Generation}\label{sec:parsing-flow-guide}
%After synthesizing the parsing map sequence $X_{parsing}$ of the target, 
The second stage aims to generate a photo-realistic foreground sequence $\mathbf{X_{fg}}$ wherein the target person dances like the person in the source video. Since the background is almost fixed, it is reasonable for the model to just focus on the foreground generation.
%We propose a Parsing-Flow Video Synthesis model for this task. 
%As the human parser above has separated the foreground person from the background of each frame, the model here only needs to generate foreground pixels of each frame which is then merged with the inpainted background to obtain the output video.
%With a slight abuse of notation, we still use $x^t$ to denote the $t$-th frame with only foreground pixels.

%Because of the less diversity of appearance in our dance dataset, 
Because of the less appearance information in one single image, it's hard for the generator to synthesize the target frame directly while keeping the appearance details. 
%Moreover, after downsampling and upsampling, most details in the appearance image will be equalized. 
Inspired by ~\cite{intrinsic-flow}, we utilize a flow regression network to calculate the appearance flow between the source and target image and use it to warp feature maps during encoding the appearance image. The warped feature will be used during upsampling and guide the generator to synthesize details as much as possible.

\noindent\textbf{Flow Regression Network.} As shown in Figure~\ref{fig:framework}, the flow regression network is a U-net architecture, which takes appearance pose $I_{pose}$ and source pose $s_{pose}^i$ as inputs, then outputs the flow map $F$ and the visibility map $V$. The flow map indicates the appearance flow between the appearance image $I_a$ and the source video frame $s^i$ while the visibility map has three discrete values indicating whether the pixel on $I_a$ is visible on $s^i$ or indicating the background respectively.

Similar to~\cite{intrinsic-flow}, we calculate the ground-truth appearance flow map and visibility map with the help of the 3D model of the source image and target image. Namely, we first fit a 3D model to both source image and target image through the state-of-the-art method~\cite{2018hmr}. Then, we project the 3D model back to the 2D image plane through the image renderer~\cite{2014opendr}. Each point on the projected image has its corresponding 3D mesh face on the same 3D model. So for each point on the target projected image, our model can either find its corresponding point on the source projected image or regard it invisible on the source projected image. Based on such relation, the appearance flow map and visibility map between the source and target projected image are calculated.
%so we can use such a relation to match points in source projected image and target projected image and then calculate the appearance flow map and visibility map. 
In our case, the source image indicates the appearance image $I_a$ while the target image indicates the source video frame $s^i$.

To train this module, we take the pose pairs calculated from the projected image pairs rather than real pose pairs as input, since the 3D model fitting process may not work correctly sometimes, which results in the mismatch between the flow maps and the real pose pairs. As for loss functions, we utilize end-point-error (EPE) loss between synthetic flow map and real flow map, and cross entropy-loss between synthesized visibility map and real visibility map.

\noindent\textbf{Parsing-Flow-Guide Generation Network.} We utilize a dual-path U-net~\cite{2015dualunet} to separately model appearance foreground and parsing map information. Different from ~\cite{intrinsic-flow}, we choose the parsing map as conditional structure information. It is because the parsing map is capable of embedding more structure information of complicated motion compared to the pose map.
%It is because our dataset includes some complicated motion while the parsing map is capable of embedding more structure information in those situations compared to the pose map.
%We show in our ablation study~\ref{sec:exp}~\xyb{ERROR} that directly using pose map as input results in poor quality of the generated image in some pose.

As illustrated in Figure~\ref{fig:framework}, the parsing encoder encodes the target parsing map $x_{parsing}^i$ generated from the first stage while the appearance encoder encodes the appearance foreground $I_a$. Each feature map in the appearance encoder is warped according to the calculated flow map $F$ and visibility map $V$. Concretely, the appearance foreground feature is warped by the flow map. Then the warped feature map is separated into visible part and invisible part according to the visibility map.
%which will be concatenated together in the next step. 
After that, these two feature maps are concatenated together and passed through two convolution layers with residual path to get final warped features. In the end, the parsing features and warped features are concatenated hierarchically which are fed into the image decoder through skip connections to generate the target foreground $x_{fg}^i$.

The loss functions used during training of this module are adversarial loss $\mathcal L_{adv}$, $L1$ reconstruction loss $\mathcal L_{L1}$ and perceptual loss $\mathcal L_{per}$~\cite{johnson2016perceptual}.
% Each loss function will be discussed in detail below.

Following \cite{wang2018vid2vid}, adversarial loss $\mathcal{L}_{adv}$ from discriminator can be given by utilizing the operator $\phi_I$:
\begin{equation}
    \begin{aligned}
        &E_{\phi_I(I_a, x_{parsing}^i, x_{fg}^i)}[logD(I_a, x_{parsing}^i, x_{fg}^i]+ \\
        &E_{\phi_I(I_a, x_{parsing}^i, \hat{x}_{fg}^i)}[log(1-D(I_a, x_{parsing}^i, \hat{x}_{fg}^i))],
    \end{aligned}
\end{equation}
where $\hat{x}_{fg}^i$ is the generated target foreground and $x_{fg}^i$ is the ground truth of it.

% We minimize the following $L1$ reconstruction loss to improve the visual quality of generated foreground:
% \begin{equation}
%     \begin{aligned}
%         \mathcal L_{L1}=||x_{fg}^i - \hat{x}_{fg}^i)||_1.
%     \end{aligned}
% \end{equation}

Inspired by~\cite{johnson2016perceptual}, we further improve the visual quality of generated foreground by minimizing perceptual loss. Specifically, we use a pretrained VGG19~\cite{Simonyan2015VeryDC} network to extract features of both the real and generated foreground, and minimizing the L1 distance between them:
\begin{equation}
\mathcal{L}_{per}(x_{fg}^i ,\hat{x}_{fg}^i)=\sum_{i=0}^{n}\lambda_i\|\phi_i(x_{fg}^i)-\phi_i(\hat{x}_{fg}^i)\|_1,
\label{eq:perceptual}
\end{equation}
in which $\phi_i(\hat{x}_{fg}^i)$ denote the \textit{i}-th feature map of the synthesized image $\hat{x}_{fg}^i$ from VGG network and $\lambda_i$ denote the weight of them.

The overall loss $\mathcal{L}$ for this stage is a weighted sum of above three loss items and $\lambda_{adv}$, $\lambda_{L1}$, $\lambda_{per}$ are weights of $\mathcal L_{adv}$, $\mathcal L_{L1}$ and $\mathcal L_{per}$ respectively.
% \begin{equation}
%     \mathcal{L}=\lambda_{adv}\mathcal{L}_{adv}+\lambda_{L1}\mathcal{L}_{L1}+\lambda_{per}\mathcal{L}_{per},
% \end{equation}
% where $\lambda_{adv}$,$\lambda_{L1}$ and $\lambda_{per}$ are coefficients of three loss items respectively.
\subsection{Spatio-Temporal Fusion Network}\label{sec:spatio-temporal}
In the third stage, we propose a spatio-temporal fusion network to fuse the synthesized foreground and background properly while keeping the temporal smoothness among the consecutive frames. Briefly, given the current synthesized foreground $x_{fg}^t$, background $bg$ and previous synthesized frame $x^{t-1}$, our model $F$ predicts the current foreground mask $fg\_mask^t$ with a ResNet-like network, which is used to fuse foreground $x_{fg}^t$ and background $bg$.
Then the third stage could be formulated as:
\begin{equation}
\begin{split}
&fg\_mask^t = F(bg, x_{fg}^t, x^{t-1})\\
&x^t = x_{fg}^t \odot fg\_mask^t + bg \odot (1 - fg\_mask^t).
\end{split}
\label{eq:fusion}
\end{equation}

More concretely, as shown in Figure~\ref{fig:framework}, at the step of synthesizing the $t$-th frame $x^t$, we first concatenate the background $bg$, $t$-th synthesis foreground $x_{fg}^t$ and previous synthesis frame $x^{t-1}$, and then feed them into a ResNet-like network which outputs a foreground mask $fg\_mask^t$. Although the foreground generated by the second stage is photo-realistic, simply overlaying the synthesized foreground and background cannot integrate the foreground and background naturally. It's straightforward to train a model generating a mask to fuse the foreground and background. Moreover, taking the previous synthesized frame $x^{t-1}$ as input forces the network to focus on the previous synthesized result, which is helpful to keep the temporal smoothness.

The background used in this stage is simply inpainted by ~\cite{2019edge-connect}, refer to section~\ref{sec:dataset} for more details. During the training of this stage, we randomly select $K$ consecutive frames in each video ($K$ is set to 5 in our experiment) to represent each video. Since there is no previous synthesized frame for the first frame, we simply ignore the synthesizing of the first frame and use the directly overlaying result of the first foreground and background as the previous synthesized frame for the second frame. The loss functions used in this stage are the $L1$ reconstruction loss $\mathcal L_{L1}$ and the perceptual loss $\mathcal L_{per}$~\cite{johnson2016perceptual} as shown in section~\ref{sec:parsing-flow-guide}.

\begin{figure*}[ht]
    \centering
    \includegraphics[width=1.0\linewidth]{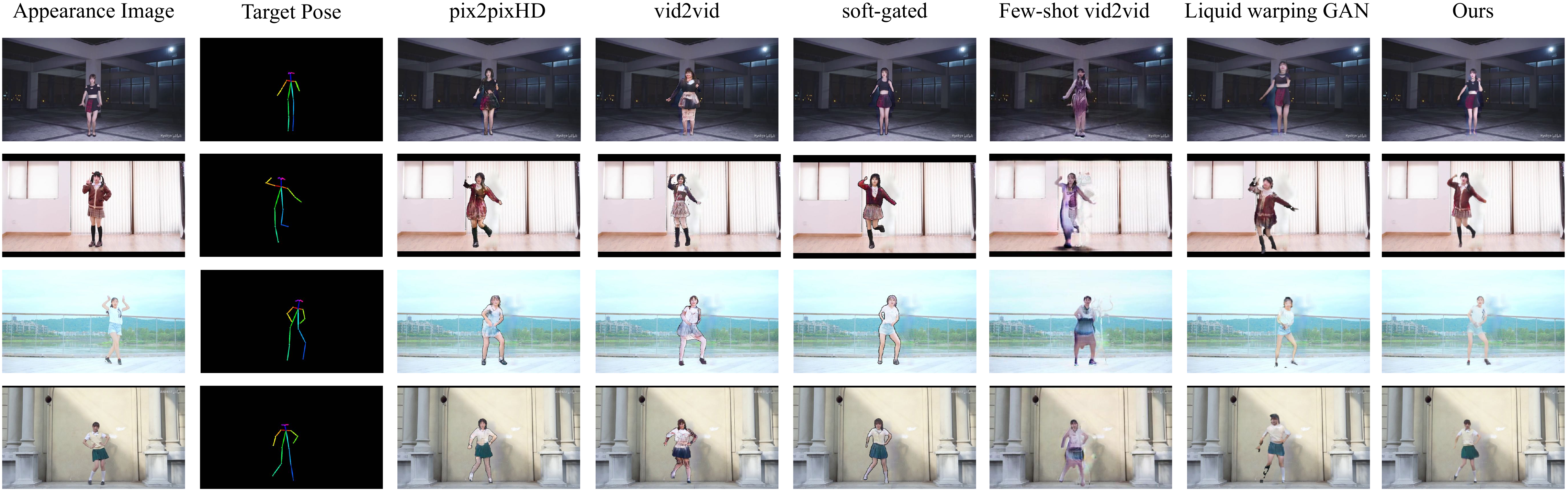}
    \caption{\revision{\textbf{Visual comparison between our method and baseline methods \wbw{on DMT dataset}.} The first column is input appearance images and the second column is the target pose we want to synthesize. Our method provides more photo-realistic results than baseline methods. Please zoom in for better view.}}
    \label{fig:fig3}
\end{figure*}

\section{Experiment}~\label{sec:exp}
%To evaluate the effectiveness of our proposed method, 
To push forward the study of human motion transfer, we construct a new large dataset for our problem (section~\ref{sec:dataset}). We compare our approach with several strong baseline models, namely pix2pixHD~\cite{wang2018pix2pixHD}, vid2vid~\cite{wang2018vid2vid} and soft-gated~\cite{soft-gated}. Both quantitative and qualitative results validate the significant superiority of our proposed approach.
\subsection{Experimental Setup}
% \noindent\textbf{Datasets}~\label{sec:dataset} For an empirical study of our video synthesis task
% The dataset used in~\cite{chan2018dance} is small and lacks appearance diversity, which is not suitable for the research field of arbitrary human motion transfer. 
% To push the limit of this research field,
\subsubsection{Datasets}
\label{sec:dataset}
We make the first attempt to propose the Dance Motion Transfer Dataset (DMT) by collecting 428 dancing videos from a video community Bilibili.
%with some sample frames shown in Figure~\ref{fig:dataset_show}. 
All of these videos satisfy the following key features: single person, almost fixed background, and great ranges of motions. The first two characteristics allow researchers to focus on the task of motion transfer instead of dealing with multiple persons or moving background. The third one makes our dataset close to real-life scenarios and makes the task challenging. %Comparing with other similar datasets, such as iPER~\cite{Liu2019LiquidWG}, DMT consists of more than ten times distinct characters and a much larger diversity of motions which brings more challenges and complexity, and also brings greater possibilities for practical application. 
We split our dataset into training and testing sets with 406 videos and 22 videos, respectively. The number of total frames for the training and testing sets are 561,785 and 31,680, respectively. The resolution of all videos is 1280x720.

Then, we adopted OpenPose~\cite{cao2018openpose,cao2017realtime,simon2017hand,wei2016cpm} to annotate pose keypoints of each frame. According to the problem setting, we also need to choose one appearance image for each video, which is conducted in the following way. Let $(x_i^j, y_i^j, c_i^j)$ be the \textit{i}-th keypoint information on the \textit{j}-th frame, where $x_i^j$, $y_i^j$ and $c_i^j$ stand for x-coordinate, y-coordinate and confidence, respectively. A single person gets $N$ keypoints on the body. In our case, $N=18$. Intuitively, an appearance image should try to show every body part of a person. We thus choose the frame with the largest confidence sum as the appearance image.
Moreover, we apply a pre-trained human parser~\cite{graphonomy} to get the parsing map of each video frame. Since the foreground occupies only a small portion of the entire image, we crop a small rectangle region containing the foreground according to the parsing map and record the coordinates of the rectangle in the original image. After padding and resize, the crop foreground size becomes $448*448$, which is the image size for the first and second stages. In the third stage, the cropped image is restored to its original size based on its cropped coordinates. Similarly, we crop the human out from the appearance image and inpaint the remaining incomplete background with EdgeConnect~\cite{2019edge-connect}. Then, the inpainted results will serve as background of each video.

Comparing with other similar datasets, such as iPER~\cite{Liu2019LiquidWG}, DMT consists of more than ten times distinct subjects and a much larger diversity of motions which brings more challenges and complexity and also brings greater possibilities for practical application. Besides, their dataset is splitted by different clothes while our dataset is splitted by different subjects.

\subsubsection{Comparison Methods}~\label{sec:comp_method}
\revision{As discussed below, we selected a wide range of methods as the baselines on this work. Pix2pixHD~\cite{wang2018pix2pixHD} and vid2vid~\cite{wang2018vid2vid} are SOTA image2image synthesis and video2video synthesis framework respectively, and the soft-gated warping GAN~\cite{soft-gated} is the latest human synthesis framework. Few-shot vid2vid~\cite{Wang2019FewshotVS} and liquid warping GAN~\cite{Liu2019LiquidWG} are latest few-shot video2video synthesis frameworks, which can synthesize videos according to single image. The implementation details of these methods are as follows.}

%\noindent\textbf{Pix2pixHD~\cite{wang2018pix2pixHD}} was proposed to solve image-to-image translation problem. We made some adaptation about pix2pixHD~\cite{wang2018pix2pixHD} so that we can apply it to our task setting. Typically, pix2pixHD~\cite{wang2018pix2pixHD} takes an image as input and also output one image. We concatenate the target pose, the appearance image and the previous $L$ generated frames $X^{t-L, t-1}$ together on channel as input to pix2pix then compute the final \textit{t}-th output frame.
%有很多没有意义又赘余的话，但是有意义需要解释清楚的又没有
\noindent\textbf{Pix2pixHD~\cite{wang2018pix2pixHD}} takes an image as input and also output one image. We made some adaptation about pix2pixHD to apply it to our task setting. Concretely, we concatenate the target pose, the appearance image and the previous $L$ generated frames $X^{t-L, t-1}$ together on the channel as input to pix2pix then compute the final \textit{t}-th output frame.

% \noindent\textbf{Vid2vid~\cite{wang2018vid2vid}} was proposed to solve video-to-video translation problem. Compared with pix2pixHD~\cite{wang2018pix2pixHD}, it considers temporal consistency and uses an extra loss to improve the visual quality of each frame on optical flow. Similar to our experiment on pix2pixHD ~\cite{wang2018pix2pixHD}, we also make some adjustments to vid2vid~\cite{wang2018vid2vid} so that it can learn conditioned on the extra appearance image. Specifically, we concatenate the appearance image with each frame of the target pose video and feed it to vid2vid~\cite{wang2018vid2vid} framework and get the resulted video.

\noindent\textbf{Vid2vid~\cite{wang2018vid2vid}}. %Compared with pix2pixHD, it considers temporal consistency and uses an extra loss to improve the visual quality of each frame on optical flow.
Similar to our experiment on pix2pixHD, % we also make some adjustments to vid2vid so that it can learn conditioned on the extra appearance image
we concatenate the appearance image with each frame of the target pose video and feed it to the vid2vid framework to get the resulted video so that it can conditioned on the extra appearance image.

% \noindent\textbf{Soft-Gated Warping GAN~\cite{soft-gated}} was proposed to solve the pose-guided person image synthesis problem. Given the target pose and appearance image, it can generate a realist person image with target pose while preserving the appearance of the target person. A frame-by-frame soft-gated warping GAN is employed as one of our baseline methods.

\noindent\textbf{Soft-Gated Warping GAN~\cite{soft-gated}} was proposed to solve the pose-guided person image synthesis problem. Given the target pose and appearance image, it can generate a realistic person image with target pose while preserving the appearance of the target person. A frame-by-frame soft-gated warping GAN is employed as one of our baseline methods.

% \revision{\noindent\textbf{Few-shot vid2vid~\cite{Wang2019FewshotVS}} was proposed to solve few-shot vid2vid synthesis problem. Here, we use the official open-source code and follow the training settings in~\cite{Wang2019FewshotVS}.}

% \revision{\noindent\textbf{Liquid Warping GAN~\cite{Liu2019LiquidWG}} is the latest few-shot person vid2vid synthesis framework. Here, we employ the official open-source code and follow the same training settings in~\cite{Wang2019FewshotVS}
% For fair comparison, we use this model to generate person region of foreground and }
\noindent\revision{\textbf{Few-shot vid2vid~\cite{Wang2019FewshotVS}}. The inputs consist of a pose sequence and some examples images. In our task setting, only one appearance frame is provided. Therefore, the appearance image is regarded as the only example image for the few-shot vid2vid model.}

\noindent\revision{\textbf{Liquid Warping GAN~\cite{Liu2019LiquidWG}}. The setting of the Liquid Warping GAN is a bit different from our task, since one of the inputs of the model is the video sequence rather than the pose sequence. Thus, during training or testing, we regard the appearance image and the target frame which corresponds to the target pose from the same video as the inputs of the model.}

% For these two methods, we both use the official open-source code and follow the official the training settings.}

\subsubsection{Implementation Details}~\label{sec:imp_detail}
% \noindent\textbf{Implementation Details}
During training, adam optimization is used for three stages with $(\beta _1=0.5, \beta _2=0.999)$.
As for the first stage, the initial learning rate is set to 0.0002. The weight for two losses in this stage is equally set to 10.0. The high-quality parsing result can be obtained within 30 epochs of training. During training the second stage, generator and discriminator are updated alternatively with batch size 8. Learning rates for generator and discriminator are initialized to 0.0002 and 0.00002 respectively. We set ${\{\lambda_{adv}, \lambda_{L1}, \lambda_{per}\}=\{0.01,1.0,1.0\}}.$ We can get realistic and natural results within 40 epochs. During training the last stage, the learning rate is initialized to 0.0001 and the batch size is set to 1 because of the high-resolution image. The loss weights are equally set to 1.0. We can obtain the natural results within 5 epochs.

%\noindent\textbf{Comparison Methods}
%We compare our approach with three baselines trained on our dataset. The detailed operation is as follows.
%Pix2pixHD~\cite{wang2018pix2pixHD} 
%We first concatenate our appearance image with target pose by channel to make pix2pixHD output images condition on appearance image. Further more, we take temporally coherence into consideration, so we also concat several previous generated frames to the input by channel.Let $P_t$, $I_a$, $S_i$ and $I_r$ denote target pose, appearance image, \textit{i}-th previous generated frames and real frame respectively. So the overall target of this baseline model is learn the mapping $F$ such that:$$F(concat([P_t, I_a, S_0...S_i]))=I_r$$
%Soft-Gated~\cite{soft-gated}
%Vid2vid~\cite{wang2018vid2vid} 

\begin{figure*}[ht]
    \centering
    \includegraphics[width=1.0\linewidth]{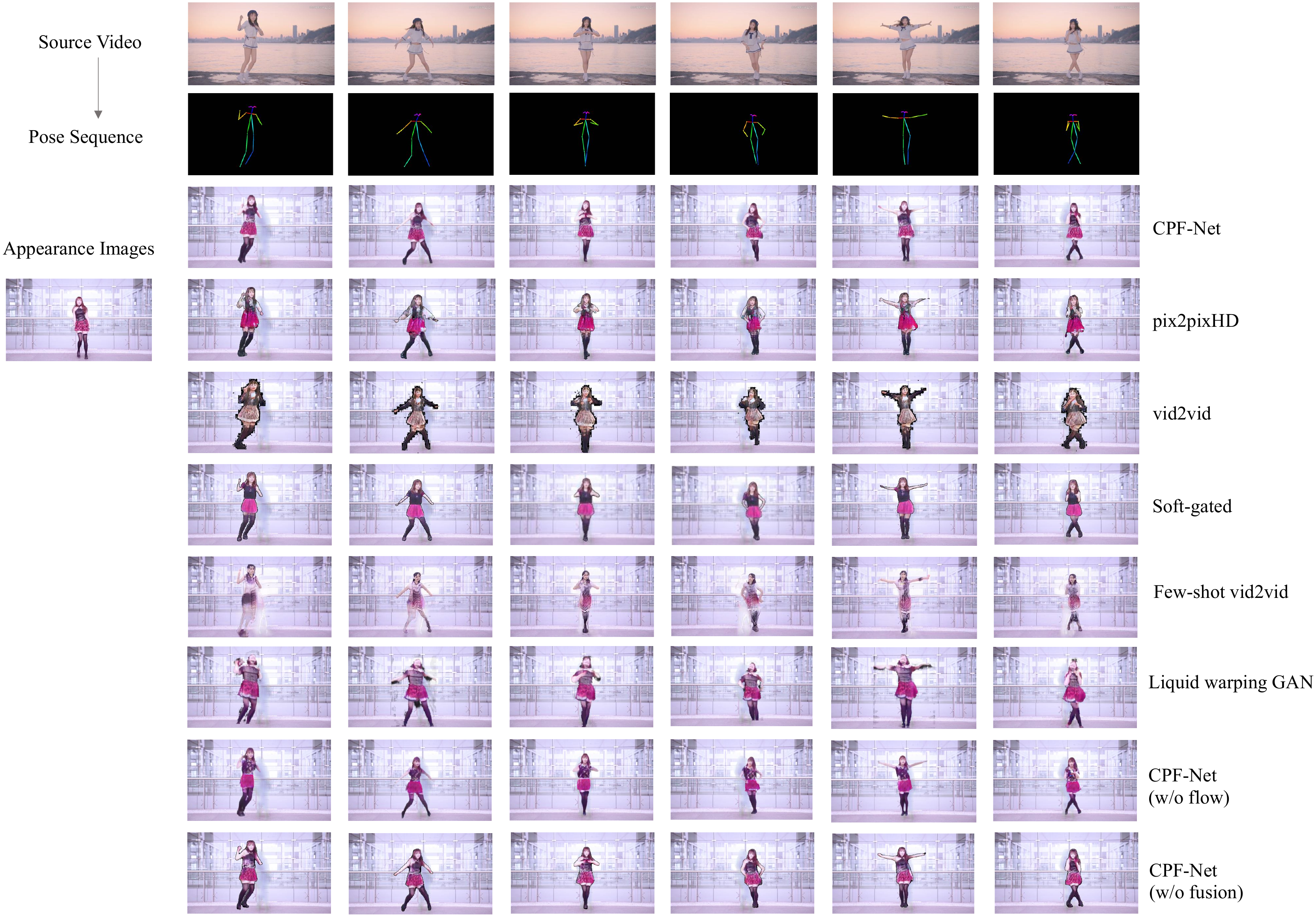}
    \caption{\revision{\textbf{Consecutive results on the test set.} The first row and second row are source video and estimated pose sequence respectively. The following rows are synthesized videos from CPF-Net, pix2pixHD, vid2vid, soft-gated, Few-shot vid2vid, Liquid warping GAN, CPF-Net(w/o appearance flow) and CPF-Net(w/o Spatio-temporal fusion) in sequence.}}
    \label{fig:fig4}
\end{figure*}

\begin{figure*}[]
    \centering
    \includegraphics[width=1.0\linewidth]{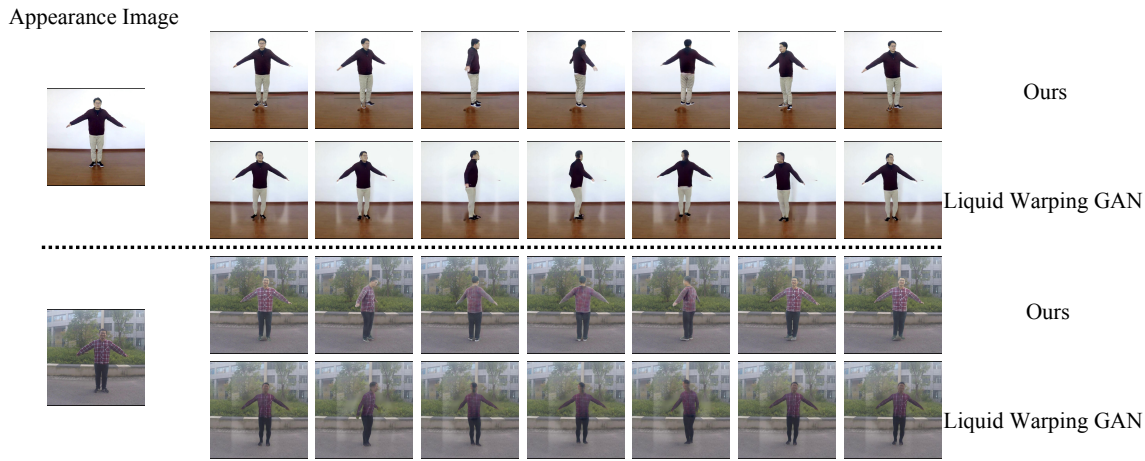}
    \caption{\wbw{\textbf{Vision comparison between our method and Liquid Warping GAN~\cite{Liu2019LiquidWG} on iPER dataset.} All models are trained with iPER dataset.}}
    \label{fig:iper}
\end{figure*}

\begin{table}[h]
\begin{center}
\begin{tabular}{|l|c|}
\hline
Method                 & Fr$\acute{e}$chet Video Distance           \\ \hline
\hline
pix2pixHD~\cite{wang2018pix2pixHD}              & 1783.94          \\ \hline
vid2vid~\cite{wang2018vid2vid}                & 2244.97          \\ \hline
soft-gated~\cite{soft-gated} & 1555.33          \\ \hline
\revision{Few-shot vid2vid~\cite{Wang2019FewshotVS}} & \revision{1608.80} \\ \hline
% \wbw{FirstOM~\cite{Siarohin2019FirstOM}        & 2823.93} \\ \hline
\revision{Liquid Warping GAN~\cite{Liu2019LiquidWG}} & \revision{2345.90} \\ \hline
\hline
Ours (w/o appearance flow)                & 1285.68          \\ \hline
Ours (w/o fusion module)                & 1454.47          \\ \hline
Ours                   & \textbf{1005.84} \\ \hline
\end{tabular}
\end{center}

\caption{\textbf{Comparison of Fr$\acute{e}$chet Video Distance between our method and baselines \wbw{on DMT dataset}.} The lower is the better.}
\label{tab:FVD_results}
\end{table}

\begin{table}[h]
\setlength{\abovecaptionskip}{5pt} 
\setlength{\belowcaptionskip}{-5pt}
\begin{center}
\begin{tabular}{|l|c|}
\hline
Comparison Method Pair & Human Eval Score \\ \hline
\hline
Ours \emph{vs} pix2pixHD~\cite{wang2018pix2pixHD}                & \textbf{0.94} \emph{vs} 0.06   \\ \hline
Ours \emph{vs} vid2vid~\cite{wang2018vid2vid}                & \textbf{0.98} vs 0.02       \\ \hline
Ours \emph{vs} soft-gated~\cite{soft-gated}  & \textbf{1.00} \emph{vs} 0.00         \\ \hline
\hline
Ours \emph{vs} Ours (w/o appearance flow)                & \textbf{0.88} vs 0.12      \\ \hline
Ours \emph{vs} Ours (w/o fusion module)  & \textbf{0.95} \emph{vs} 0.05       \\ \hline
\hline
Ours (F) \emph{vs} pix2pixHD (F)~\cite{wang2018pix2pixHD}  &      \textbf{0.93} \emph{vs} 0.07       \\ \hline
Ours (F) \emph{vs} vid2vid (F)~\cite{wang2018vid2vid}  &     \textbf{0.96} \emph{vs} 0.04    \\ \hline
Ours (F) \emph{vs} soft-gated (F)~\cite{soft-gated}   &    \textbf{0.88} \emph{vs} 0.12  \\ \hline
\end{tabular}
\end{center}

\caption{\textbf{Human evaluation results comparing pairs of methods \wbw{on DMT dataset}.}(F) means that only generated foreground is used for evaluation.}
\label{tab:Human}
\end{table}

\begin{table}[h]
\begin{center}
\begin{tabular}{|l|c|}
\hline
Method   & Fr$\acute{e}$chet Video Distance           \\ \hline
Liquid Warping GAN~\cite{Liu2019LiquidWG}  &  1134.48 \\ \hline

Ours    & \textbf{1083.62} \\ \hline
\end{tabular}
\end{center}
\caption{\wbw{\textbf{Comparison of Fr$\acute{e}$chet Video Distance between our method and Liquid Warping GAN~\cite{Liu2019LiquidWG} on iPER dataset.} The lower is the better.}}
\label{tab:FVD_results_iper}
\end{table}

\subsection{Quantitative Results}
\textbf{Fr$\mathbf{\acute{e}}$chet Video Distance (FVD)}~\cite{unterthiner2018towards} is a new metric for generative models of video based on Fr$\acute{e}$chet Inception Distance (FID)~\cite{Heusel2017GANsTB}. This metric considers both temporal consistency and visual quality of each frame with the lower score representing the better result. ~\cite{unterthiner2018towards} conducted a large-scale human study and confirmed that FVD correlates well with qualitative human judgment of those generated videos. On our evaluation, the network used to calculate FVD is I3D~\cite{2017i3d}.
%Formally, we choose resulted from frames $\{R_i\}_{i=L_{min}/2}^{L_{min}/2+K}$ of each generated video as the sequences to evaluate FVD, where $L_{min}$ be the minimum length of all videos in the test set.
We calculate FVD between source videos of test set and generated videos. Corresponding to the released code of FVD, we selected 30 consecutive frames of each video to calculate the FVD.
%We used our trained model and baselines to generate videos in the testing set. 
% Let $I_a$, $S=\{R_i\}_{i=0}^{N-1}$ and $S^{'}=\{R_i^{'}\}_{i=0}^{N-1}$ be appearance image, source video in testing set and the generated video in the testing set. Then we calculate FVD between $\{S\}$ and $S^{'}$. Corresponding to the released code of FVD, we selected 30 consecutive frames of each video to calculate the FVD.

Table~\ref{tab:FVD_results} shows the comparison results among our method and the other three baselines. We can observe that our method gets the lowest FVD score.
% The FVD score of soft-gated~\cite{soft-gated}, pix2pixHD~\cite{wang2018pix2pixHD} and vid2vid~\cite{wang2018vid2vid} ranks in an increasing order.
\revision{The FVD score of soft-gated~\cite{soft-gated}, Few-shot vid2vid~\cite{Wang2019FewshotVS}, Liquid Warping GAN~\cite{Liu2019LiquidWG}, pix2pixHD~\cite{wang2018pix2pixHD} and vid2vid~\cite{wang2018vid2vid} ranks in increasing order.}
%We can observe that vid2vid~\cite{wang2018vid2vid}, pix2pixHD~\cite{wang2018pix2pixHD}, soft-gated~\cite{soft-gated} and our method achieve smaller distance in order.
The videos generated by vid2vid~\cite{wang2018vid2vid} get the largest distance among all methods. Pix2pixHD~\cite{wang2018pix2pixHD} and soft-gated~\cite{soft-gated} focus more on synthesis of single frame while vid2vid~\cite{wang2018vid2vid} pays more attention to the fluency of video, which may cause degradation on generated videos of vid2vid~\cite{wang2018vid2vid}. \wbw{Tabel~\ref{tab:FVD_results_iper} shows the comparison results among our method and Liquid Warping GAN~\cite{Liu2019LiquidWG} on iPER dataset. We can observe that our method gets the lower FVD score than Liquid Warping GAN, and this illustrates that our method is of good generalization.}
% \revision{Liquid Warping GAN~\cite{Liu2019LiquidWG} and few-shot vid2vid~\cite{Wang2019FewshotVS}  achieve relatively better result among all baseline methods, since they are proposed for few-shot video2video synthesis problem.}

% In fact, FVD samples part of the consecutive video frames and computes the distribution distance with ground-truth video.

%Table~\ref{tab:FVD_results} shows the results of FVD~\cite{unterthiner2018towards}. We can observe that vid2vid~\cite{wang2018vid2vid}, pix2pixHD~\cite{wang2018pix2pixHD}, vid2vid~\cite{wang2018vid2vid} (no background) and our method achieve smaller distance in order. Videos generated by vid2vid~\cite{wang2018vid2vid} got larger distance than pix2pixHD~\cite{wang2018pix2pixHD}. In fact, FVD samples part of the consecutive video frames and computes the distribution distance with ground-truth video. Pix2pixHD~\cite{wang2018pix2pixHD} focuses more on synthesis of single frame while vid2vid~\cite{wang2018vid2vid} pays more attention to the fluency of video, which may cause degradation on generated videos of vid2vid~\cite{wang2018vid2vid}. In addition, we also find that methods with background processed separately including vid2vid~\cite{wang2018vid2vid} (no background) and our method acquired better metric results. For one thing, such a model can be more concentrated on the learning of pose transfer to realize better results without consideration on the background. For another, the inpainted background may be more likely to resemble the background of the target video.

\begin{figure*}[ht]
    \centering
    \includegraphics[width=0.9\linewidth]{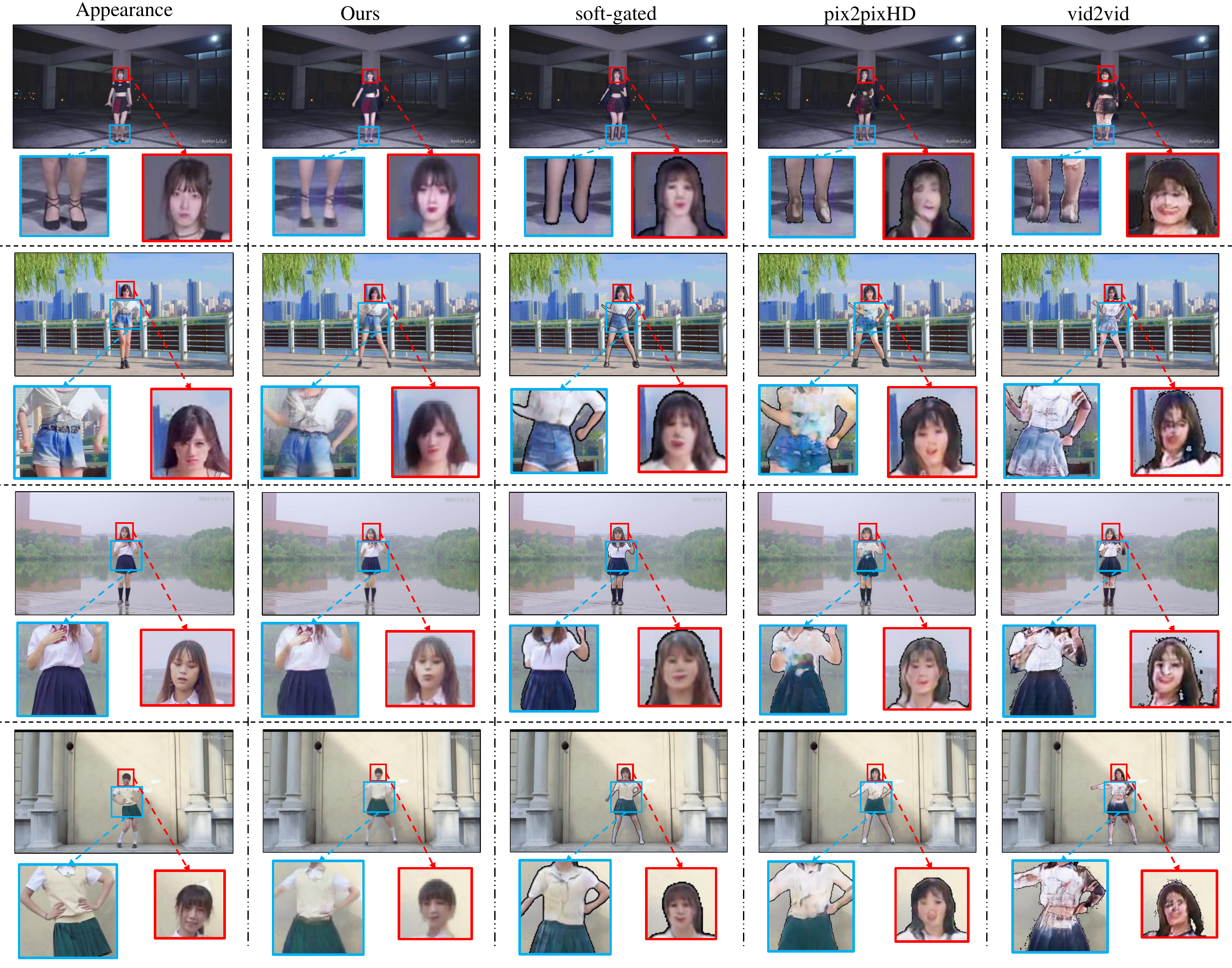}
    \caption{\textbf{Detail comparison between our method and baseline methods.} The images on the first column are input appearance images and the images lies on the second column are generated by our method and the images lies on the remaining columns are generated by baseline methods. The enlarged detail images illustrate that CPF-Net is capable to generate much more face details(e.g.,  cleaner hair)and much better clothes texture than baseline methods.}
    \label{fig:detail}
\end{figure*}

\subsection{Human Evaluation}
For video synthesis tasks, human evaluation is always the most accurate and reliable. 
%Once human cannot tell the difference of the real video and the fake one, video synthesis achieves the goal. In fact, there is still large gap between reality and the goal. 
We conduct our human evaluation on Amazon Mechanical Turk. In each test, we provide two videos(about 10 seconds) to each participant, in which one comes from the compared method(baseline method or ablation method) and the other comes from our proposed method. The expectation for them is to choose the more realistic and natural video from the two. All videos of the test set are used for evaluation and each video will be evaluated by $5$ independent workers. To ensure the accuracy of test results, we require the workers to be master qualified by Amazon.

The human evaluation results among our method and the other three baselines are demonstrated in Table~\ref{tab:Human}. Our method shows absolute superiority in authenticity and naturalness by ratio 0.94, 0.98, 1.00 compared to baselines pix2pixHD~\cite{wang2018pix2pixHD}, vid2vid~\cite{wang2018vid2vid} and soft-gated~\cite{soft-gated}. To further evaluate our proposed method on foreground generation, human evaluation with foreground videos was conducted.Our approach still outperforms with the three baseline methods with 0.93, 0.96 and 0.88 on foreground generation respectively. To sum up, videos generated by our model have significantly better visual quality from the perspective of a human.

\subsection{Qualitative Results}
We first compare the motion transfer results of baselines and our method in the single frame, as shown in Figure ~\ref{fig:fig3}. Obviously, our method acquires better results both on clothes detail retention and human outline. The visual qualities of pix2pixHD~\cite{wang2018pix2pixHD} and vid2vid~\cite{wang2018vid2vid} are poorer than the other three methods. This owes to the simple synthesizing from the generator, resulting in the lack of appearance details and some wrong appearance synthesis. Human images generated by soft-gated~\cite{soft-gated} and liquid warping GAN~\cite{Liu2019LiquidWG} seem better than the other three baselines due to their warping operation. However, our method still outperforms soft-gated, where the enlarged images in Figure~\ref{fig:detail} shows our method maintains better facial details as well as clothing texture.
% indicating the appearance flow is capable of guiding the synthesis of appearance details.
%However, our method still outperforms soft-gated, indicating the appearance flow is capable of guiding the synthesis of appearance details.
%People in results of Pix2pixHD~\cite{wang2018pix2pixHD} and vid2vid~\cite{wang2018vid2vid} are severely distorted. 
%This owes to the processing of background and foreground at the same time, causing less concentration on the generation of poses and the appearances of the dancing human, which can be verified by the comparison of vid2vid~\cite{wang2018vid2vid} and vid2vid~\cite{wang2018vid2vid} (no background). 
Additionally, we also provide the consecutive results of generated videos of our method and baselines in Figure~\ref{fig:fig4}. Similar to the single frame, our method performs better in transferring poses and retaining the details of appearance. \wbw{Figure~\ref{fig:iper} shows the visual comparison between our proposed method and liquid warping GAN~\cite{Liu2019LiquidWG} on iPER dataset. Our proposed method achieved better visual quality than liquid warping GAN, and this verified the generality of our proposed method.}

% \begin{figure}[ht]
%     \centering
%     \includegraphics[width=0.9\linewidth]{dance/fig/comparison.pdf}
%     \caption{\textbf{Detail comparison between our method and soft-gated.}The images lies on the left column are generated by proposed CPF-Net while those on the right column are generated by soft-gated~\cite{soft-gated}. The enlarged detail images illustrate that CPF-Net is capable to generate much more face details(e.g., cleaner hair and choker on the neck) and much better clothes texture(e.g., plaid on the skirt, off-the-shoulder t-shirt and love pattern on clothes) than soft-gated.}
%     \label{fig:detail}
% \end{figure}

%To demonstrate the internal effectiveness of our methods, we present the intermediate parsing results and final synthesized images, as displayed in Figure ~\ref{fig:test}. These results show that we generated reasonable human motion parsings and those parsings contribute a lot to the synthesis of realistic-look video.

\subsection{Ablation Study}
We conduct two additional experiments to verify the effectiveness of different modules in CPF-Net, namely CPF-Net without using appearance flow and CPF-Net without spatio-temporal fusion. The implementation details of these ablation methods are as follows.

\textbf{CPF-Net(w/o flow).} Flow Regression Network of second stage is not used. The feature map extracted from the appearance foreground $I_a$ is directly fed into the image decoder without feature warping.

\textbf{CPF-Net(w/o fusion).} Mask $fg_{mask}^t$ generated by Spatio-Temporal Fusion Network is not used. Generated parsing map $x_{parsing}^i$ is directly used as mask to fuse foreground and background.

The  results of FVD are shown in Table~\ref{tab:FVD_results},
%, while the results of human evaluation are shown in tabel~\ref{tab:Human}. 
while some visual results are presented in Figure~\ref{fig:fig4}. The full model of CPF-Net obtains the best score of FVD compared to the other two methods, which indicates each module of proposed CPF-Net is essential for the synthesized video. %our proposed method can improve the visual quality and temporal consistency of the synthesized video.
Moreover, the FVD score of CPF-Net without spatio-temporal fusion module still outperforms the baselines, which demonstrates the visual quality of the single frame generated by the second stage is better than the other baselines. As shown in Figure~\ref{fig:fig4}, the synthesized frames generated by CPF-Net without using appearance flow usually lack of appearance details. CPF-Net without spatio-temporal fusion module can not integrate the foreground and background naturally, which usually generates a boundary between the foreground and background. Moreover, Table~\ref{tab:Human} shows that videos generated by the full model of CPF-Net gains a much higher score on human evaluation than ablation models which means much better visual quality is gained by full model. To sum up, both subjective and objective metric show the necessity of each part of CPF-Net.

% \begin{figure}[]
%     \setlength{\abovecaptionskip}{5pt}
%     \setlength{\belowcaptionskip}{-5pt}
%     \centering
%     \includegraphics[width=0.9\linewidth]{fig/comparison_one.pdf}
%     \caption{\textbf{Detail comparison between our method and soft-gated.}The images lies on the left and right column are generated by our method and soft-gated~\cite{soft-gated} respectively. The enlarged detail images illustrate that CPF-Net is capable to generate much more face details(e.g., cleaner hair) and much better clothes texture(e.g., plaid on the skirt and neckline shape) than soft-gated.}
%     \label{fig:detail}
% \end{figure}

\section{Conclusion}
In this work, we introduce a more general motion transferring task where we aim to learn a single model to parsimoniously transfer motion from a source video to any target person given only one image of the target. We propose a structured approach, named as Collaborative Parsing-Flow Video Synthesis (CPF-Net), to guide the synthesis of photo-realistic foreground, which will be merged into the background by a spatio-temporal fusion module.
%which integrates target human parsing and appearance flow between source and target human to guide the realistic foreground synthesis and generate the temporal consistent video through a Spatio-temporal fusion module.
%with flow-guided video synthesis.
The problem is decoupled into stages of human parsing generation, photo-realistic foreground generation, and temporal consistent video generation. We demonstrate the integration of human parsing map and appearance flow bring a large improvement to the visual quality in the final video while the spatio-temporal fusion module guarantees the natural fusion of foreground and background and the temporal smoothness of the synthesis video. \revision{We conduct comparison experiments with a wide range of baseline methods, both quantitative results and qualitative results demonstrate the superiority of our method.}

%We demonstrated that introducing structured human parsing map sequence as the intermediate results brings a large improvement when synthesizing video from motion sequence. The impressive performance compared to other baselines indicate the effectiveness of the proposed method for this challenging motions transferring task.

%Limitations and future work. Although our approach outperforms previous methods, our model still has some limitations. For example, our model fails to synthesize a realistic face of the target for lacking specialized face GAN to increase facial realism. Furthermore, our model relies on the quality of synthesized human parsing to some extent, which would be incorrect sometimes. This might be resolved if we utilized the pose map as prior information when generating the foreground.

% simply overlaying the synthesized foreground and the inpainted background still can not integrate the foreground and the background naturally. This might be resolved if we train a generative model to merge the foreground and background.

\section*{Acknowledgments}

\wbw{This work was supported in part by National Key R\&D Program of China under Grant No. 2020AAA0109700, National Natural Science Foundation of China (NSFC) under Grant No.U19A2073 and No.61976233, Guangdong Province Basic and Applied Basic Research (Regional Joint Fund-Key) Grant No.2019B1515120039, Guangdong Outstanding Youth Fund (Grant No. 2021B1515020061), Shenzhen Fundamental Research Program (Project No. RCYX20200714114642083, No. JCYJ20190807154211365), CSIG Youth Fund.}

% \fi
\IEEEpeerreviewmaketitle

\ifCLASSOPTIONcaptionsoff
  \newpage
\fi



% \end{thebibliography}



%\vfill

% \clearpage
\bibliographystyle{plain}
\bibliography{egbib}

\begin{IEEEbiography}[{\includegraphics[width=1in,height=1.25in,clip,keepaspectratio]{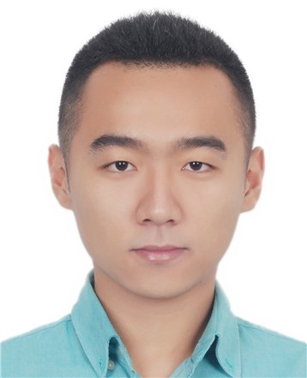}}]{Bowen Wu}
received his BE degree and is currently working toward the ME degree in the School of Data and Computer Science, Sun Yat-sen University, China. His research interests include generative model and its applications.
\end{IEEEbiography}

\begin{IEEEbiography}[{\includegraphics[width=1in,height=1.25in,clip,keepaspectratio]{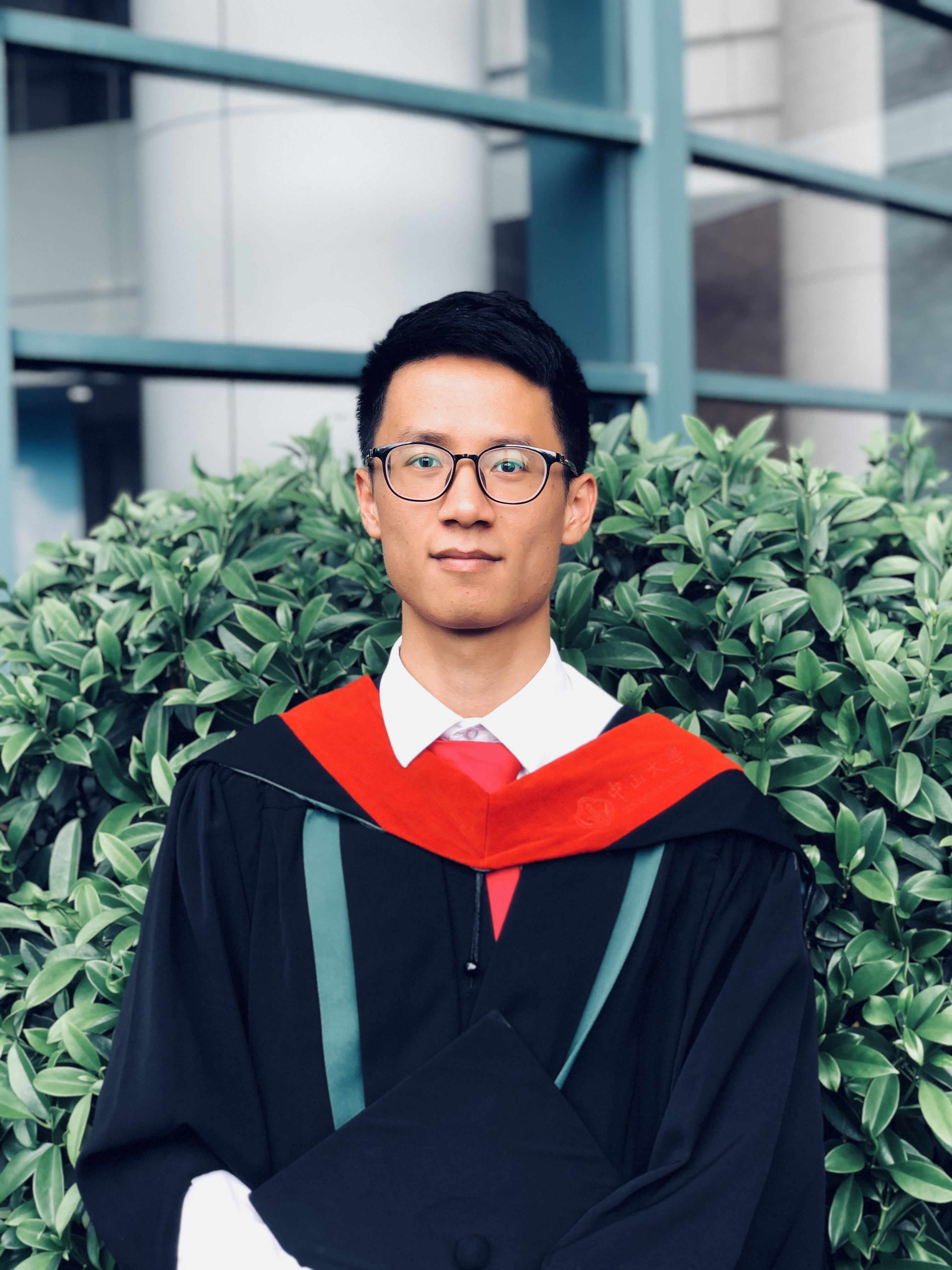}}]{Zhenyu Xie} received his bachelor degree from Sun Yat-sen University and is studying for the master degree in the School of Data and Computer Science, Sun Yat-sen University. Currently, his research interests consist of image synthesis and video synthesis.

\end{IEEEbiography}

\begin{IEEEbiography}[{\includegraphics[width=1in,height=1.25in,clip,keepaspectratio]{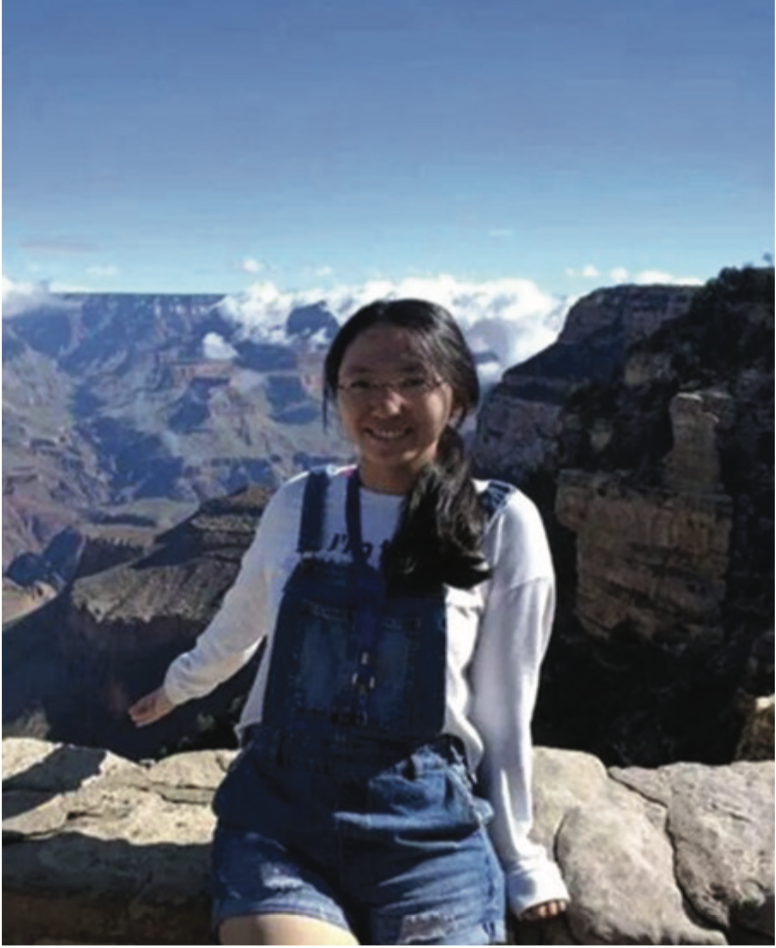}}]{Xiaodan Liang}
is currently an Associate Professor at Sun Yat-sen University. She was a postdoc researcher in the machine learning department at Carnegie Mellon University, working with Prof. Eric Xing, from 2016 to 2018. She received her PhD degree from Sun Yat-sen University in 2016, advised by Liang Lin. She has published several cutting-edge projects on human-related analysis, including human parsing, pedestrian detection and instance segmentation, 2D/3D human pose estimation and activity recognition.
\end{IEEEbiography}

\begin{IEEEbiography}[{\includegraphics[width=1in,height=1.25in,clip,keepaspectratio]{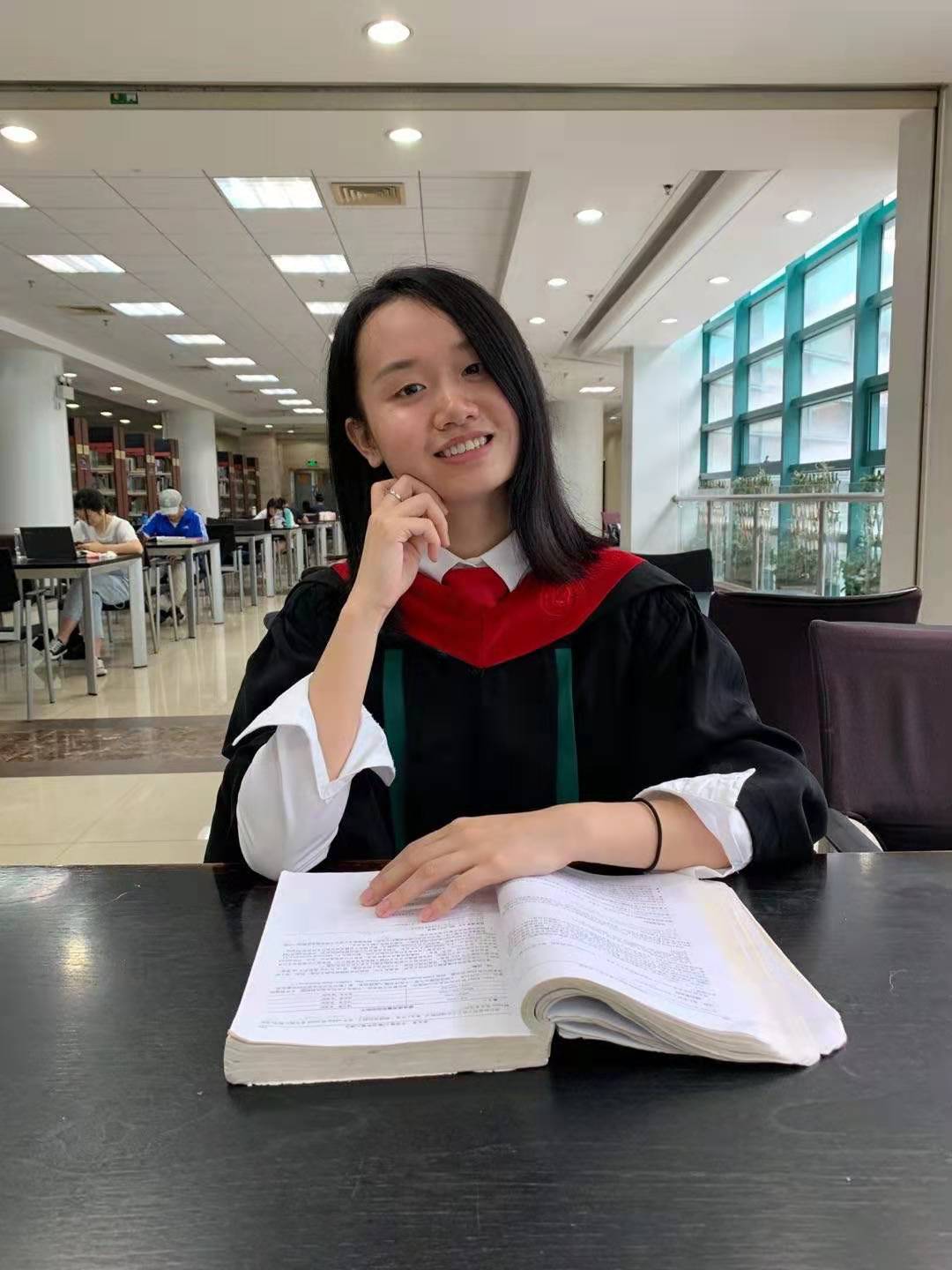}}]{Yubei Xiao}
received her BE degree and is currently working toward the ME degree in the School of Data and Computer Science, Sun Yat-sen University, China. Her research interests include unsupervised learning and meta-learning for speech recognition.
\end{IEEEbiography}

\begin{IEEEbiography}[{\includegraphics[width=1in,height=1.25in,clip,keepaspectratio]{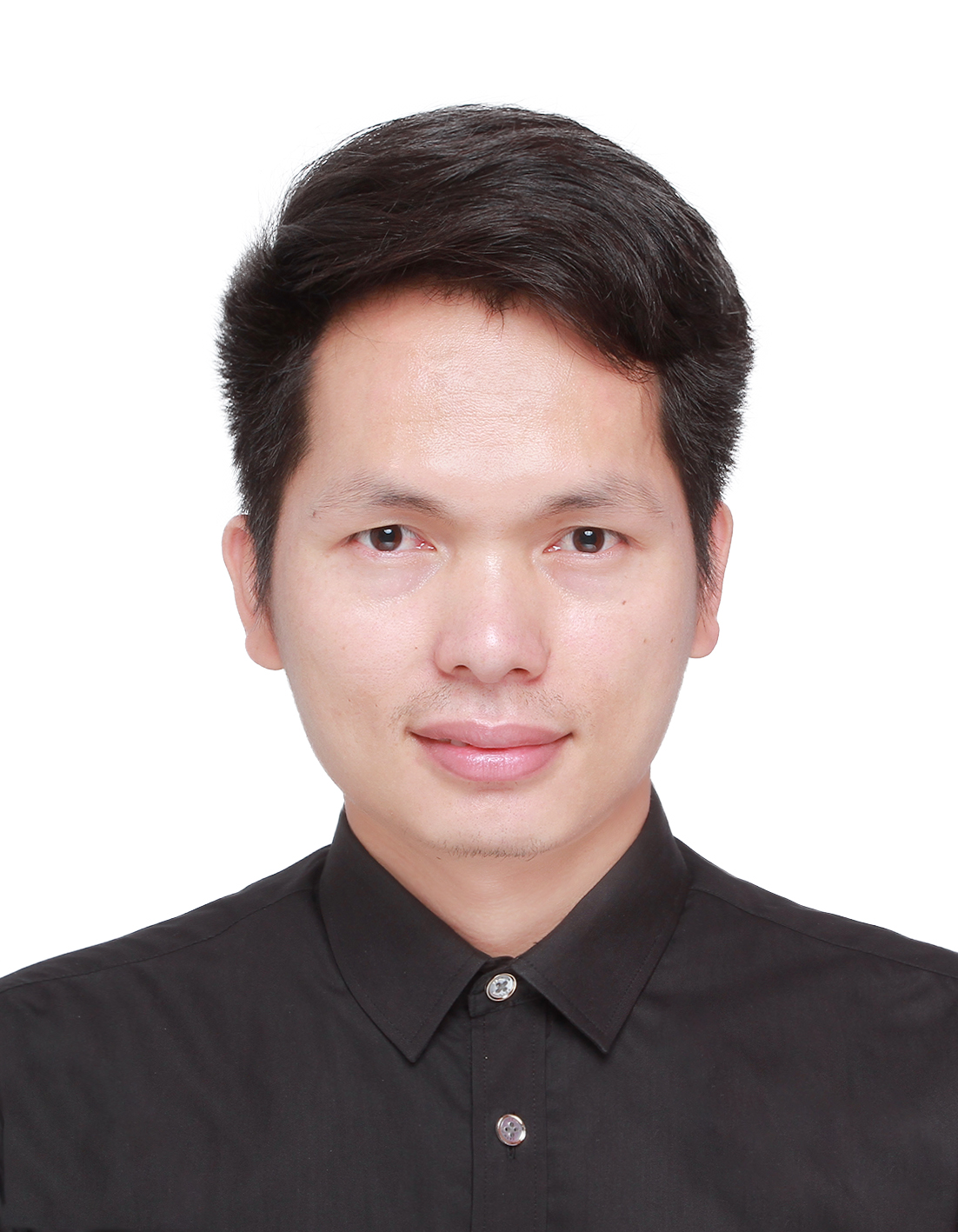}}]{Haoye Dong}
is a Ph.D student at School of Data and Computer Science, Sun Yat-sen University, advised by Prof. Jian Yin. He received his master degree from School of Computer Science, South China Normal University in 2016, advised by Prof. Yong Tang. He received his bachelor degree from  School of Computer Science, Guangdong Polytechnic Normal University in 2012. His research interests are deep generative learning and human-centric image/video synthesis.
\end{IEEEbiography}

% \begin{IEEEbiography}[{\includegraphics[width=1in,height=1.25in,clip,keepaspectratio]{}}]{Qixian Zhou}
% Biography text
% \end{IEEEbiography}

% \begin{IEEEbiography}[{\includegraphics[width=1in,height=1.25in,clip,keepaspectratio]{}}]{Bingcheng Chen}
% Biography text
% \end{IEEEbiography}

\begin{IEEEbiography}[{\includegraphics[width=1in,height=1.25in,clip,keepaspectratio]{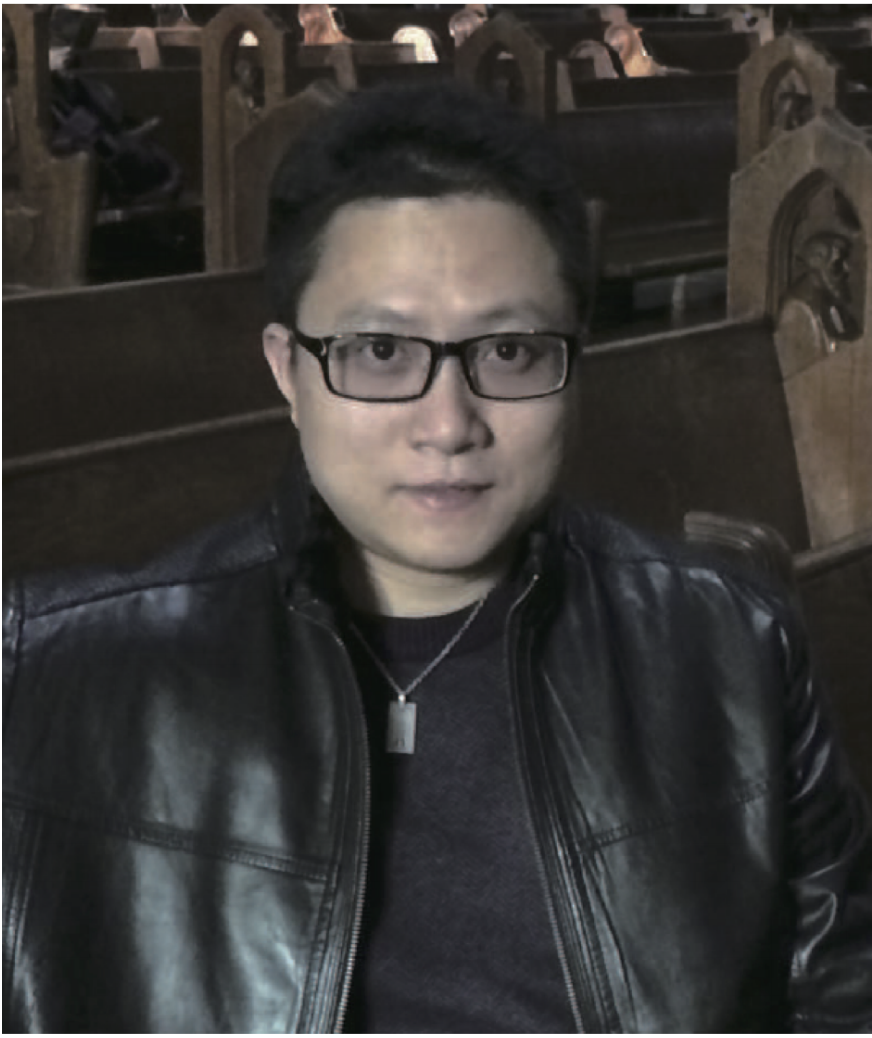}}]{Liang Lin}
is CEO of DMAI Great China and a full professor of Computer Science in Sun Yat-sen University. He served as the Executive Director of the SenseTime Group from 2016 to 2018, leading the R\&D teams in developing cutting-edge, deliverable solutions in computer vision, data analysis and mining, and intelligent robotic systems. He has authored or co-authored more than 200 papers in leading academic journals and conferences (e.g.,TPAMI/IJCV, CVPR/ICCV/NIPS/ICML/AAAI). He is an associate editor of IEEE Trans, Human-Machine Systems and IET Computer Vision, and he served as the area/session chair for numerous conferences, such as CVPR, ICME, ICCV, ICMR. He was the recipient of Annual Best Paper Award by Pattern Recognition (Elsevier) in 2018, Dimond Award for best paper in IEEE ICME in 2017, ACM NPAR Best Paper Runners-Up Award in 2010, Google Faculty Award in 2012, award for the best student paper in IEEE ICME in 2014, and Hong Kong Scholars Award in 2014. He is a Fellow of IET.
\end{IEEEbiography}

%\newpage
\end{document}